\documentclass[runningheads]{llncs}
\usepackage{graphicx}
\usepackage{amsmath,amssymb} 
\usepackage{floatrow}
\usepackage[font=scriptsize,skip=0pt]{caption}
\usepackage{subcaption}
\usepackage{color}
\usepackage{textcomp}
\usepackage[export]{adjustbox}
\usepackage{wrapfig,lipsum,booktabs}
\usepackage{wrapfig}
\usepackage{verbatim}

\usepackage{import}
\newcommand\vs{7}

\begin{document}

\mainmatter

\title{OOWL500: Overcoming Dataset Collection Bias in the Wild}



\author{Brandon Leung, Chih-Hui Ho, Amir Persekian, David Orozco, Yen Chang, Erik Sandstrom, Bo Liu, Nuno Vasconcelos}
\institute{UC San Diego \\
\email{\{b7leung, chh279, aperseki, dorozco, yec084, esandstr, boliu, nvasconcelos\}@ucsd.edu}
}

\maketitle

\begin{abstract}
The hypothesis that image datasets gathered online ``in the wild"  can produce biased object recognizers, e.g. preferring professional photography or certain viewing angles, is studied.  A new ``in the lab'' data collection infrastructure is proposed consisting of a drone which captures images as it circles around objects. Crucially, the control provided by this setup and the natural camera shake inherent to flight mitigate many biases. It's inexpensive and easily replicable nature may also potentially lead to a scalable data collection effort by the vision community. The procedure's usefulness is demonstrated by creating a dataset of {\it Objects Obtained With fLight\/} (OOWL). Denoted as OOWL500, it contains $120,000$ images of $500$ objects  and is the largest ``in the lab" image dataset available when both number of classes and objects per class are considered. Furthermore, it has enabled several of new insights on object recognition.  First, a novel adversarial attack strategy is proposed, where image perturbations are defined in terms of semantic properties such as camera shake and pose. Indeed, experiments  have shown that ImageNet has considerable amounts of pose and professional photography bias. Second, it is used to show that the augmentation of in the wild datasets, such as ImageNet, with in the lab data, such as OOWL500, can significantly decrease these biases, leading to object recognizers of improved generalization. Third, the dataset is used to study questions on ``best procedures'' for dataset collection. It is revealed that data augmentation with synthetic images does not suffice to eliminate in the wild datasets biases, and that camera shake and pose diversity play a more important role in object recognition robustness than previously thought.
\end{abstract}

\section{Introduction}
The last few years have shown that large convolutional neural networks (CNNs), such as Alexnet~\cite{krizhevsky2012imagenet}, VGG~\cite{simonyan2014very}, GoogleNet~\cite{szegedy2015going},
ResNet~\cite{he2016deep}, etc. achieve previously unmatched object recognition
performance.  It is also well established that these models can be
easily transferred to other tasks by fine-tuning. As a result, CNNs are now
almost universally used in computer vision. Many variants on
object recognition architectures have also been proposed for object
detection~\cite{girshick2014rich,ren2015faster,he2017mask}, action recognition~\cite{simonyan2014two,tran2015learning},
image captioning~\cite{karpathy2015deep}, question answering~\cite{antol2015vqa}, etc.
In summary, object recognition breakthroughs have had a surprisingly
large ``multiplier effect'' for the whole of computer vision.

Besides CNNs, much of this progress can be ascribed to the introduction of ImageNet ~\cite{deng2009imagenet}. 
With 1 million images covering 1,000 object
classes, it is one of the largest datasets in the literature. Additionally, the ease with which ImageNet trained networks can be
transfered to other tasks suggests that it enables the learning of
{\it universal visual representations\/}. It could be 
argued that, at least for  lower convolutional layers of modern CNNs,
filters learned on ImageNet are applicable to most vision
tasks. While fine-tuning on specific data may enable gains,
these are usually marginal. In many cases, the simple use of
ImageNet CNNs as ``feature extractors''  is a strong baseline. All of
this suggests that ImageNet experiments enable fundamental conclusions about
object recognition. For example, because modern CNNs achieve smaller error
than human annotators on ImageNet, it has been claimed, both in  technical~\cite{Pneumonia2017paper,HeZR015,AIindex2017report} and
popular~\cite{Pneumonia2017news,Nervana2015news,Microsoft2015news} literature, that CNNs
have superhuman object recognition performance. 

\begin{figure*}[t]\RawFloats
  \centering
  \includegraphics[width=1\linewidth]{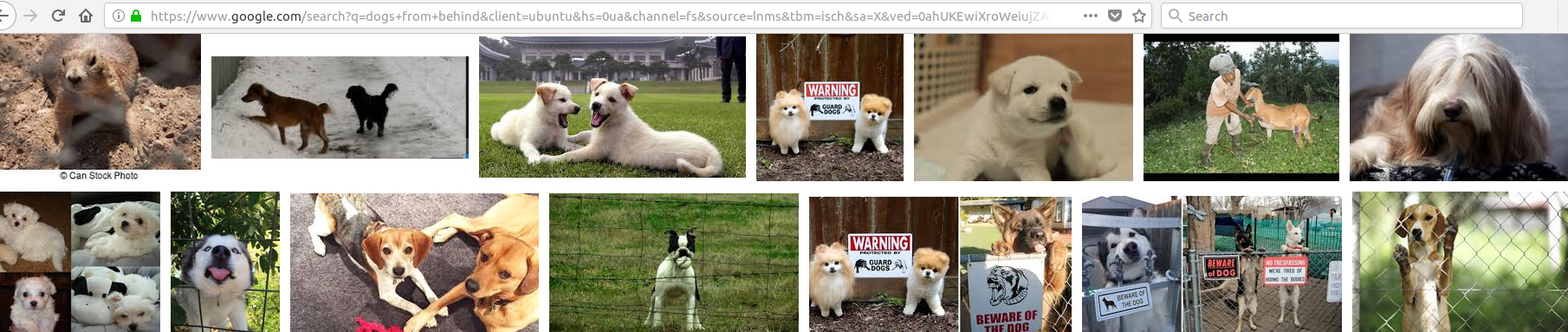}
  \vspace{1pt}
  \caption{Results of a Google image query for {\it dogs from behind}.}
  \label{fig:dfb}
\end{figure*}
	
Nevertheless, experience has shown that most datasets have biases~\cite{Torralba2011bias}. 
Like many recent vision datasets, ImageNet is the product of data
collection ``in the wild'' -- images are first collected on the
internet, then annotated on platforms like MTurk. Dataset collection 
in the wild is a major departure from classic vision practice of
collection ``in the lab''. Its main advantage is removing dataset biases ensuing from a limited number of object classes, objects
per class, lighting patterns, backgrounds, imaging systems, etc.
However, there are at least two significant limitations. First, images
collected on the web have some biases of their own. This is because
 people who upload do a certain amount of self selection, usually
seeking to publish ``good'' images. Consequently, it is not easy to collect large quantities of images of objects in the wild under {\it unpopular imaging
  settings\/}, such as ``poor'' lighting, ``uncommon'' or ``improper''
object views, ``poor'' focus, motion blur, etc. These unpopular settings
vary with the object class. While trees are photographed from all angles
and distances, dog pictures tend to depict close-ups of dog faces.
Fig. \ref{fig:dfb} illustrates this point by showing a set
of images returned by Google image search for the query
{\it ``dogs from behind''\/}. This makes it likely that object recognizers
learned in the wild will have much more difficulty recognizing dogs from
behind than frontally.
	
A second important limitation is that the lack of explicit control
over the imaging process makes difficult to determine ImageNet's biases. In fact, because the images used to evaluate performance are
also collected in the wild, this can be impossible. A lack of
``dogs from behind'' in the test set makes it impossible to know if a
recognizer fails on such images. Even when the biases can be observed,
it is nearly impossible to quantify them. Since Turkers are poor estimators 
of variables such as object viewing angle (which we denote as {\it pose\/}), lighting angle, etc.,
it is difficult to recover these annotations a posteriori. This
hampers research in topics such as pose-invariant
or light-invariant object recognition. We believe that progress along these
directions requires bringing dataset collection back ``to the lab'' and enabling explicit
control over variables such as object pose, distance, or lighting direction.
Several previous
works have addressed these problems~\cite{nenecolumbia,geusebroek2005amsterdam,lecun2004learning}. 
However, many of these efforts predate deep learning, aiming for
datasets much smaller and less diverse than what is expected
today. Camera domes or turntables were favored, which can be expensive and cumbersome
to build and maintain. We believe that more scalable data collection
procedures, which can quickly be disseminated and replicated throughout the
vision community, are needed to address today's requirements.

\begin{figure}[t]\RawFloats
%
  \begin{minipage}{0.9\linewidth}
    \centering
    \begin{tabular}{cccc}
    	\centering
    	\includegraphics[width=0.24\textwidth]{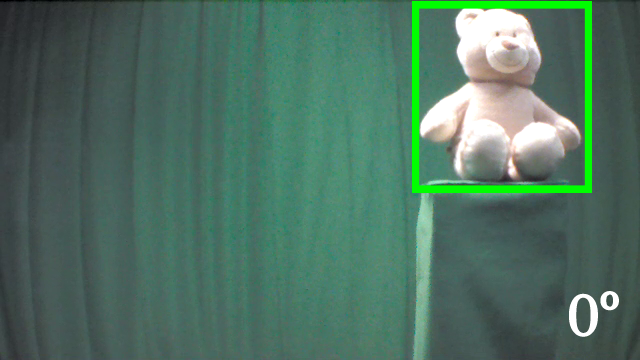}
        & \includegraphics[width=0.24\textwidth]{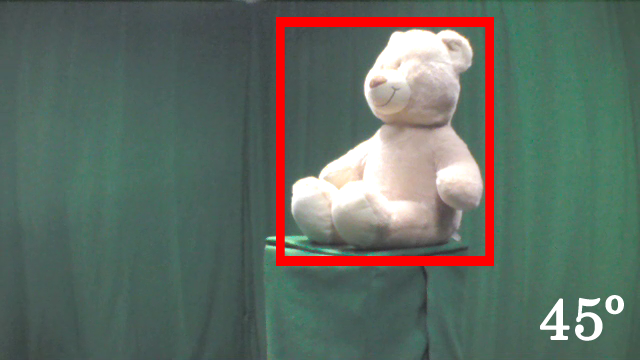}
        & \includegraphics[width=0.24\textwidth]{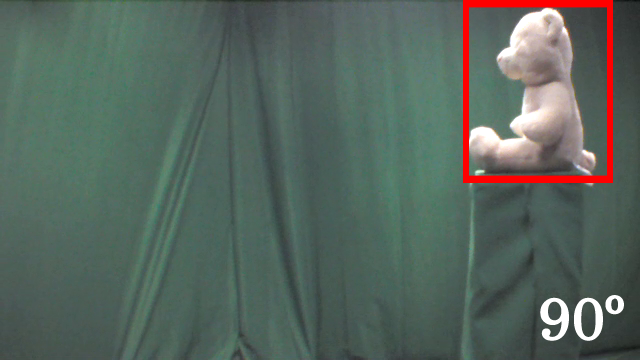}
        & \includegraphics[width=0.24\textwidth]{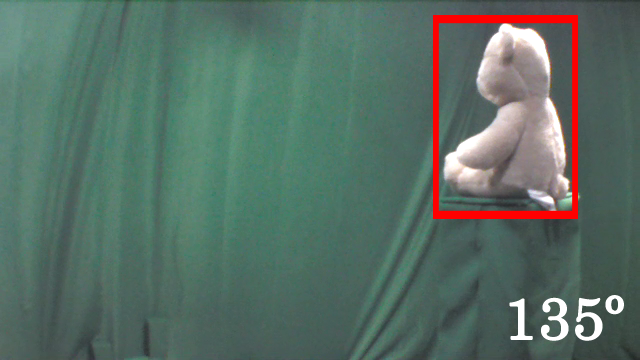}\\
        
        \includegraphics[width=0.24\textwidth]{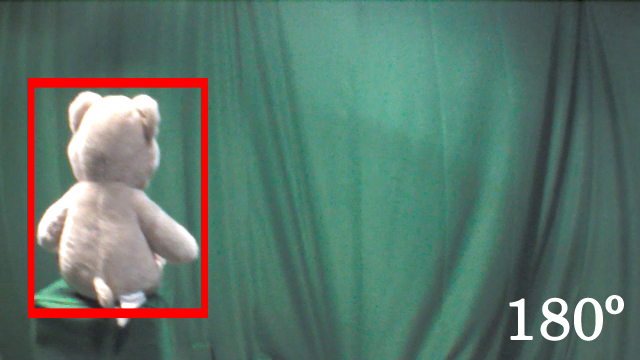}
        & \includegraphics[width=0.24\textwidth]{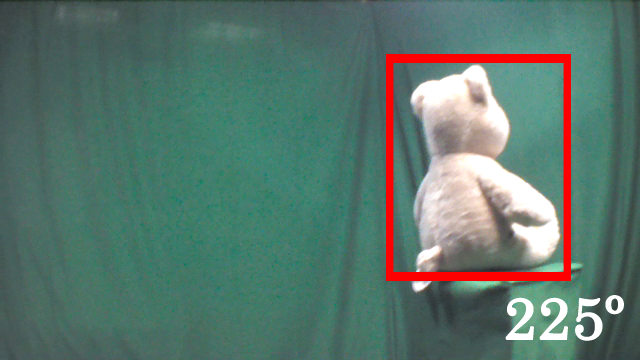}
        & \includegraphics[width=0.24\textwidth]{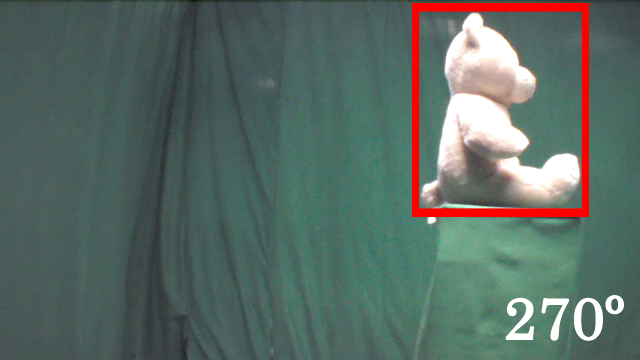}
        & \includegraphics[width=0.24\textwidth]{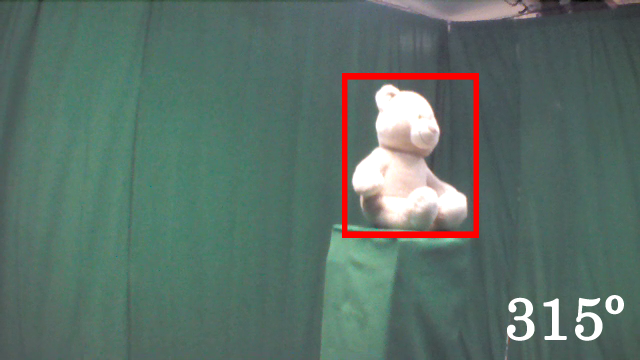}\\
    \end{tabular}\\
    \scriptsize{(a)}
  \end{minipage}
   \begin{minipage}{0.078\linewidth}
    \centering
    \begin{tabular}{c}
  		\includegraphics[width=\textwidth]{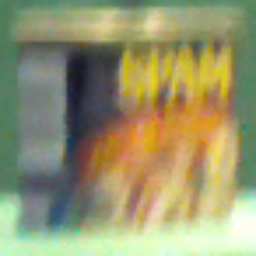} \\
  		\includegraphics[width=\textwidth]{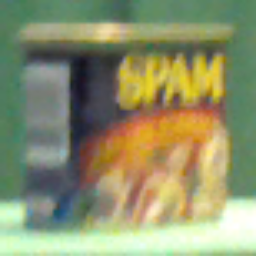} \\
  		\includegraphics[width=\textwidth]{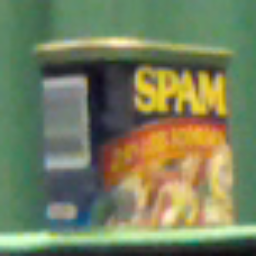} \\
    \end{tabular}
    \scriptsize{(b)} 
  \end{minipage}
 
  \caption{(a) Example images collected  per viewing angle with bounding boxes.
    Green highlights the frontal view. (b) Examples of varying levels of focus captured while hovering.}
  \label{fig:bears}
\end{figure}

\setlength\intextsep{0pt}
\begin{wrapfigure}{r}{0.28\textwidth}
	\centering
	\includegraphics[width=\linewidth]{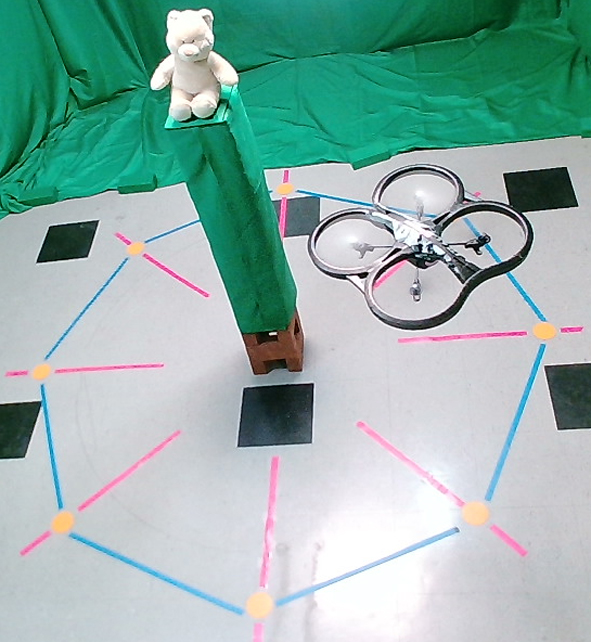}
	\caption{The proposed drone based data collection setup.}
	\label{fig:setup}
	\vspace{-5pt}
\end{wrapfigure}
In this work, we address this problem by proposing a new solution to
scalable dataset collection ``in the lab''. This leverages the fact that
drones are becoming ubiquitous, cheap, and easy to program, allowing
the collection of controlled object datasets to be scaled up. We consider the problem of pose invariant recognition,
and propose a drone based setup to collect images over a set of object
views. As illustrated in Fig. \ref{fig:setup} this consists of
having a drone fly around the object and collecting pictures at
precise pose intervals. The inexpensive drone selected is commercially available, 
 and the setup only requires tape and software. Hence, it 
 can be easily replicated or scaled to different environments.
Since all code will be made publicly available, we believe that
it could enable a community wide effort on in the lab data collection.
This is not easy with turntables or camera domes, which
are more expensive and substantially harder to build
and maintain. We also demonstrate the feasibility of the proposed setup by introducing a new dataset of
{\it Objects Obtained With fLight\/} (OOWL). Denoted
OOWL500, it contains $120,000$ images of $500$ different objects, imaged from
$8$ viewing angles. The images are labeled for both identity (object
and object class), pose, and
other metadata that facilitates a number of experiments discussed in the
paper.

OOWL500 has several unique properties compared to past pose datasets.
First, it is larger in regards to its combination of number of classes and number of
object instances per class. Second, unlike datasets collected with domes
or turntables,
there is a small ``camera shake'' effect that produces
images with statistics different than those of ImageNet style datasets.
While still perfectly recognizable to humans, motion blur and somewhat lower resolutions pose additional
difficulties to ImageNet trained classifiers, beyond the lack of pose
robustness. This is an effect similar to adversarial attacks that are
now commonly studied in the literature~\cite{papernot2017cleverhans,kurakin2016adversarial}. Hence, OOWL500 is a stronger challenge for ImageNet classifiers than previous pose
datasets. Third, OOWL500 has larger class overlaps with ImageNet than previous datasets and 
was designed to additionally overlap with a synthetic pose dataset, ModelNet~\cite{wu20153d}.
All of this enables several experiments that were not possible with previous datasets.

We leverage all these properties to investigate numerous questions
regarding the robustness of current object recognizers. A number of
surprising observations ensue from these experiments. These include
the fact that ImageNet classifiers have far weaker generalization than
humans, that it is quite easy to mount adversarial attacks on current
classifiers by simply manipulating object pose, that pose diversity is
as important as instance diversity for training object recognizers,
that synthetic data is not very helpful in mitigating the limitations of datasets collected in the wild, and that the
augmentation of ImageNet with OOWL500 style images leads to more
robust recognition. While some of these conclusions are not
qualitatively surprising, the {\it strength\/} of the
effects, i.e. the gains or losses in recognition accuracy, can be
surprisingly large. This shows that much more research is needed in object
recognition methods with good generalization across poses and higher
robustness in general. We thus believe that the combination of our
experimental observations, and the introduction of OOWL500, will inspire
substantial further research in this area.

\vspace*{-\vs pt}
\section{The OOWL500 dataset} \label{dataset}
\vspace*{-\vs pt}
In this section, we present the proposed drone-based setup for 
image collection ``in the lab'' and the resulting OOWL500 dataset.

\vspace*{-\vs pt}
\subsection{Dataset collection setup}
\vspace*{-\vs pt}
Fig. ~\ref{fig:setup} shows the proposed setup for 
image collection ``in the lab''. A drone flies around an object, following
flight commands issued by a state machine implemented in
software which relies on colored markings on the 
floor for navigation.
The drone always takes off from the same 3D location and stops at orange
circles on the ground. At these locations, it faces the object
to take pictures at particular viewing angles, producing the images shown in Fig. ~\ref{fig:bears}(a).
The enclosure and  the platform used to place objects upon are surrounded
entirely by green screens, preventing recognition algorithms from
exploiting background cues. Since the objects are ultimately cropped, using the bounding boxes shown in
Fig. ~\ref{fig:bears}(a), there are no strong background effects.

The flight space is illuminated by six LED lamps attached to the  ceiling, ensuring approximately
homogeneous illumination of the object under all views. As shown
in Fig. ~\ref{fig:bears}(a), there can still be lighting variations on
the object surface. This is due to surface curvature or
other properties, not the existence of a preferred light direction.
To prevent object recognition algorithms from exploiting lighting
patterns to ascertain pose, the placement of objects' front face was randomized in regards to which wall it faced.
The procedure used to control the drone flight is facilitated by its two cameras.
One faces the ground, and is used for navigation. The second faces
the object, and is used to collect object views. 
Images collected by the ground facing camera are processed by
standard OpenCV vision operators to detect the floor markings.
As the drone approaches the object, a PID controller issues commands to adjust its altitude, center the orange circle in the ground view, and rotate towards the pink line's direction. Once in place, the drone hovers and takes 30 pictures using the frontal camera at a rate of 5 Hz.
At this point, the drone uses the ground camera to follow the blue line to the next orange circle. This sequence of commands loops until the drone finishes flying around the object. The software is implemented so that new states can easily be programmed and inserted into the sequence in any order, allowing for many potential flight configurations.

Crucially, no further intervention is needed once the drone has been placed on the first stop. A single button press instructs the drone to calibrate, take off, fly around the object, save images, and land autonomously in about 5 minutes. 
A video of a typical flight is provided as supplemental material,
and precise details about the setup as well as all code
will be released upon publication of the paper. This
will make the infrastructure easily reusable, as a black box, by anyone who intends
to generate a dataset or augment OOWL500. Furthermore, the code is written to accommodate flight paths of any reasonable size and shape, as indicated by the floor markings.
Because only a laptop, a drone, and colored tape suffice to capture objects of many different sizes within a wide range of backgrounds and lighting conditions, we hope that these factors will result in a community wide effort to generate a very large dataset.



Another interesting facet that we plan to explore in the future is that
 although this is an ``in the lab'' setup, it can actually be deployed
outside the lab. This raises the possibility of a mixed ``in the lab'' and
``in the wild'' setting, where the procedure is deployed outdoors against
natural backgrounds but retains the ability to control variables like pose
and distance to the object. The setup can even be scaled up
to very large objects such as entire buildings or monuments. Currently, the choice to optimize the setup for low-end drones
prevents us from doing this. We have so far used a Parrot A.R. Drone quadcopter because, at a
cost of about \$200, it comes equipped with an ultrasound altimeter, an
accelerometer, a gyroscope, and two (forward and bottom facing) cameras used 
to facilitate data collection.  This was primarily
motivated by the desire of a very low barrier to entry for those interested in
contributing to dataset collection. This is unlike most previous ``in the lab''
image collection procedures, based on turntables
or domes, that are much more expensive and difficult to build.
However, the current drone can be  challenging to control outdoors. We plan to support other 
drones in the future.
\vspace*{-\vs pt}	
\subsection{Dataset}
\vspace*{-\vs pt}
The above setup was used to create the OOWL500 dataset, which
contains $500$ objects captured at varying angles, for a total of $120,000$ images.
The objects are evenly distributed among $25$ classes
of everyday objects, such as backpacks, bottles, shoes, and teddy bears.
To facilitate the experiments in Section \ref{experiments}, these classes
have been chosen to maximize overlap with those of
ImageNet~\cite{deng2009imagenet} and ModelNet~\cite{wu20153d}.
There are $20$ objects per class, imaged from the $8$
views (poses) of Fig.~\ref{fig:setup}. Since $30$ images are taken
per view, each object originates $240$ images, for a total of $120,000$ images. 
As shown in Fig. \ref{fig:bears}(a), views are
acquired in steps of 45\textdegree. The $240$ pictures are  stored as
$640 \times 360$ PNGs.

\begin{figure}[t]
  \begin{minipage}{0.47 \linewidth}
    \centering
    \def\w{0.15\textwidth}
    \scriptsize
    \begin{tabular}{cccc}
      \includegraphics[width=\w, height=\w]{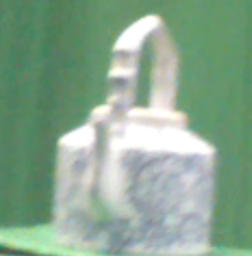}
      & \includegraphics[width=\w, height=\w]{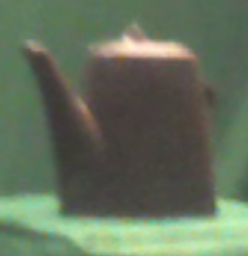}
      & \includegraphics[width=\w, height=\w]{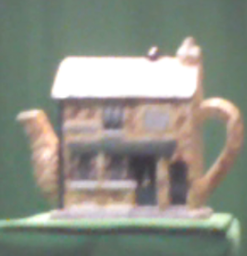} 
      & \includegraphics[width=\w, height=\w]{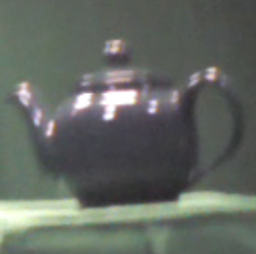} \\
      $0^{\circ}$ & $45^{\circ}$ & $90^{\circ}$ & $135^{\circ}$\\
      \includegraphics[width=\w, height=\w]{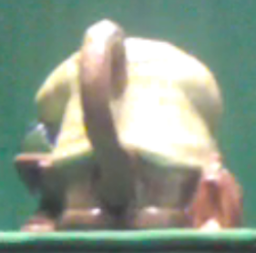}
      & \includegraphics[width=\w, height=\w]{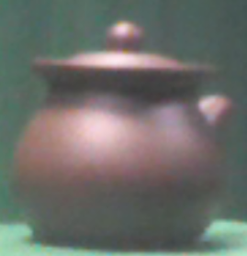} 
      & \includegraphics[width=\w, height=\w]{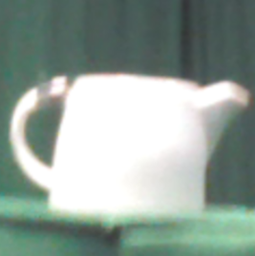}
      & \includegraphics[width=\w, height=\w]{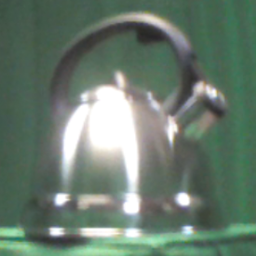} \\
      $180^{\circ}$ & $225^{\circ}$ & $270^{\circ}$ & $315^{\circ}$
    \end{tabular}
    \hspace{.2in}\\
    \scriptsize{(a)} 
  \end{minipage} \hspace{.1in}
  \begin{minipage}{0.48 \linewidth}
    \centering
    \includegraphics[width=0.7\textwidth]{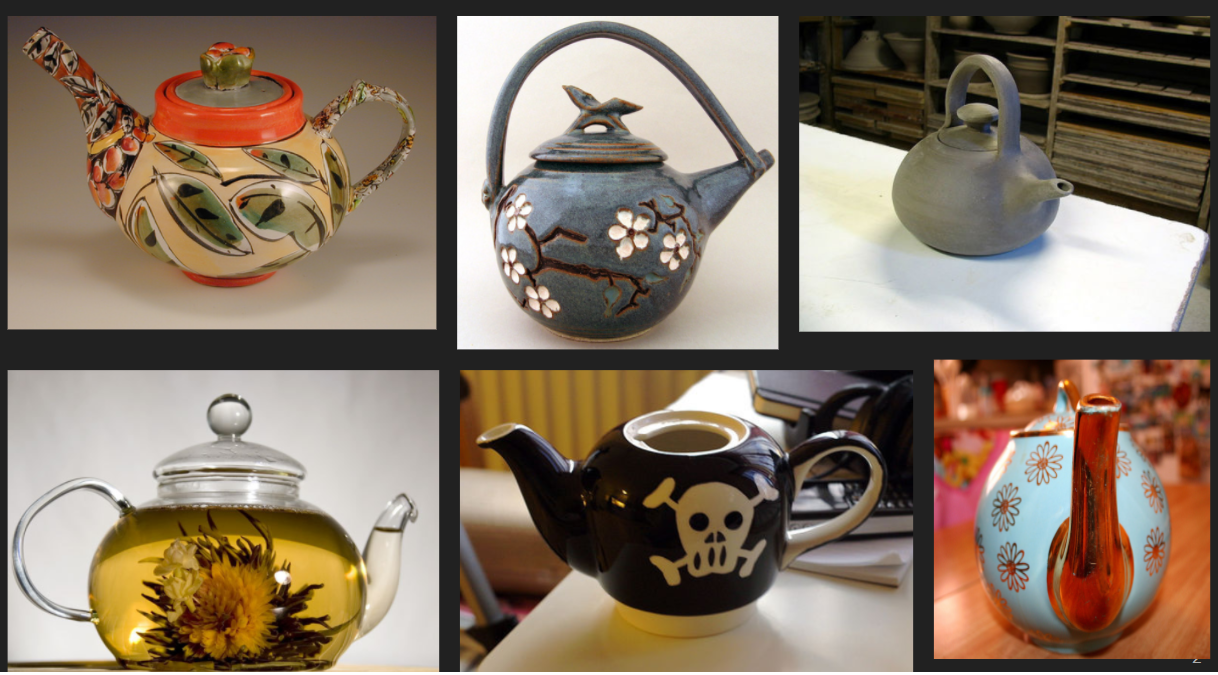} \\
    \vspace{.07in}
    \scriptsize{(b)}
  \end{minipage} \vspace{-.1in}
  \caption{Images from the ``teapot'' object class. Left: OOWL500.
    Right: ImageNet.}
  \label{fig:egimages}
\end{figure}

Besides the richness of poses, OOWL500 has an adversarial aspect
for classifiers trained on previous datasets, due to the somewhat unorthodox
nature of images collected by the drone. While several pictures portray the same object at the same pose, they vary in object position, blurriness,
exposure, and noise. Shown in Fig. \ref{fig:bears}(b), this is due to camera shake that ensues as the drone
hovers in place. Some images in the ``teapot'' class are shown in Fig.~\ref{fig:egimages}(a).
Note the different ``feel'' of these images vs. teapot examples from ImageNet, shown in Fig. ~\ref{fig:egimages}(b). The latter illustrate an ImageNet bias towards ``professional quality'' object shots, which results from the ``self-selection'' usually performed for images posted on the web. The camera shake effect of OOWL500
can be thought as a generalization of the standard manipulations, such as
pixel-swapping, commonly used in adversarial attacks to
CNNs~\cite{papernot2017cleverhans,kurakin2016adversarial} or image shifting and rescaling commonly used to augment
training sets. The variation due to hovering naturally generates images
for either attacking existing networks or augmenting
training sets to build better ones. This trait is explored further in
Section \ref{subsec:ImageNet bias}.

Each image in the dataset is accompanied by a CSV
file that  provides the  following annotated information regarding the
object: a  short description; the pose in degrees relative to
other pictures of that object; whether or not the picture depicts the
object's frontal face (denoted by a 1 or 0 respectively, or a -1 if the
object does not have an easily identifiable front face);  the object's
degree of rotational symmetry (indicated by the next degree increment
where the object appears identical); and coordinate pairs specifying the
bounding box. 
\begin{table}[t!]
  \begin{scriptsize}
    \centering
    \begin{tabular}{|l|ccccccc|}
      \hline
      Dataset & \parbox[t]{1.2cm}{Domain} & \parbox[t]{1.2cm}{Classes} & \parbox[t]{1.2cm}{Objects \\per cls} & \parbox[t]{1.2cm}{Images \\per obj} & \parbox[t]{1.2cm}{Total \\Images} & \parbox[t]{1.2cm}{Overlap\\ImageNet} & \parbox[t]{1.2cm}{Overlap\\OOWL} \\
      \hline \hline
      & \multicolumn{7}{c|}{In the wild} \\
      \hline \hline
      Caltech-101~\cite{fei2007learning} & General & 102 & 90 & 1 & 9,144 & & 7 \\
      Pascal~\cite{everingham2010pascal} & General & 20 & 790-10,129 & 1 & 11,540 & & 5 \\
      Caltech-256~\cite{griffin2007caltech} & General & 257 & 119 & 1 & 30,607 & & 17 \\
      \hline
      MNIST~\cite{lecun1998gradient}  & Digit & 10 & 6,000 & 1 & 60,000 & & 0 \\
      CIFAR-100~\cite{krizhevsky2009learning}  & General & 100 & 600 & 1 & 60,000 & & 9 \\
      MS COCO~\cite{lin2014microsoft}  & General & 80 & $\sim$11,079 & 1 & 123,287 & & 17 \\
      ImageNet~\cite{deng2009imagenet} & General & 21,841 & $\sim$650 & 1 & 14,197,122 & & 23\\
      \hline \hline
      & \multicolumn{7}{c|}{In the lab} \\
      \hline \hline
      Table-Top-Pose~\cite{sun2010depth} & Table Top & 3 & 10 & $\sim$16 & 480 & 2 & 1 \\
      NYC3DCars~\cite{matzen2013nyc3dcars} & Car & 1 & 3,787 & 1 & 1,287 & 1 & 1 \\
      EPFL Car~\cite{ozuysal2009pose} & Car & 1 & 20 & $\sim$115 & 2,299 & 1 & 1 \\
      ETH-80~\cite{leibe2003analyzing} & General & 8 &  10 & 41 & 3,280 & 4 & 1 \\
      Multi-PIE~\cite{gross2010multi} & Face & 1 & 337 & 15 & 5,055 & 0 & 0 \\
      COIL~\cite{nenecolumbia} & General & 100 & 1 & 72 & 7,200 & - & - \\
      FERET~\cite{phillips2000feret} & Face & 1 & 1199 & 1-24 & 14,051 & 0 & 0 \\
      PASCAL 3D+~\cite{xiang2014beyond} & General & 12 & $\sim$1158 & $\sim$3 & 30,899 & 10 & 6 \\
      NORB~\cite{lecun2004learning} & Toy & 5 & 10 & 1944 & 48,600 & 1 & 3 \\
      T-LESS~\cite{hodan2017t} & Industry & 30 & 1 & 648-1,296 & $\sim$49,000 & 0 & 0  \\
      \hline
      ALOI~\cite{geusebroek2005amsterdam} & General & 1000 & 1 & 72 & 72,000 & - & 15 \\
      iLab-20M~\cite{borji2016ilab} & Toy Vehicle & 15 & 25-160 & 1,320 & 21,798,480 & 11& 4 \\
      HM~\cite{lai2011large} & General & 51 & $\sim$6 & $\sim$750 & 250,000 & 19 & 7 \\
      OOWL500 & General & 25 & 20 & 240 & 120,000 & 23 & - \\
      \hline \hline
      & \multicolumn{7}{c|}{Synthetic} \\
      \hline \hline
      ModelNet~\cite{wu20153d} & General & 40 & 100 & 3D & 3D & 28 & 10 \\
      ShapeNet-55~\cite{shapenet2015} & General & 55 & $\sim$940 & 3D & 3D & 46 & 16 \\
      \hline
    \end{tabular}
    \caption{Overview of well known object recognition datasets. For datasets with pose annotations, the number of overlapped classes with ILSVRC2012 are listed. This is not applicable for COIL and ALOI however (only individual objects are provided instead of classes).}
    \label{tab:dataset}
  \end{scriptsize}
\end{table}

\subsection{Relation to previous datasets} \label{relatedWork}
\vspace*{-\vs pt}
\subsubsection{In the wild datasets}
Table~\ref{tab:dataset} compares OOWL500 to
previous datasets. The top third of the table summarizes the
properties of ``in the wild'' datasets.  These will
continue to have a critical role in object recognition; 
they cover a diversity of imaging scenarios, objects, scenes, and
scene configurations not replicable in the lab. 
They are listed mostly for calibration of what constitutes a large object recognition dataset today.  
Note that, with less than $30,000$ images, Pascal and the two Caltech datasets are now considered ``small'' in comparison to MS-COCO and ImageNet. At $60,000$ images, MNIST and CIFAR are still widely used in machine learning, but less so in computer vision. This suggests a threshold of $60,000$ images, marked by the horizontal lines in the table.

The goal of this work is not to replace in the wild datasets. Rather, we seek to complement  them by investigating two of their limitations.  The first is that they exhibit some {\it biases\/} resulting from image collection
on the web. One example is the {\it professional
photography\/} bias of Fig. ~\ref{fig:egimages}(b). Since most people post
images to convey a message, they often carefully self-select the images to post.
Hence, objects can have ``standardized poses'' that makes them more easily recognizable or appealing. 
This helps explain the {\it pose bias\/} of Fig. \ref{fig:dfb}. This bias is also visible
on the ImageNet teapots of Fig.~\ref{fig:egimages}(b), which tend to
be presented from a side view. The second limitation of ``in the wild'' datasets is the lack of
information needed to study important questions in object recognition,
such as invariance to pose, scale, lighting, etc.
Consider pose variability. Under mild assumptions on
 imaging setup (e.g. uniform lighting), a rigid object spans a
3D manifold in image space as the viewing angle changes. This suggests a
decomposition into the {\it appearance} and
the {\it pose} information of the object. An object recognizer could,
in principle, exploit this decomposition to improve its
generalization, e.g. by leveraging a model of pose trajectories learned from
some objects to better recognize others, for which less training views are
available. This is easier if the dataset contains {\it pose
  trajectories}, i.e. multiple views of each object, and even more likely
if the dataset is annotated with {\it pose labels}. None of the two hold
for ``in the wild'' datasets, which favor instance diversity. For
example, ImageNet contains 1,000 images of chairs but
no different views of the same chair. Even if this were the
case, pose labels would be difficult to recover on MTurk.

\setlength\intextsep{5pt}
\begin{wrapfigure}{r}{0.63\textwidth}
  \centering
  \includegraphics[width=\linewidth]{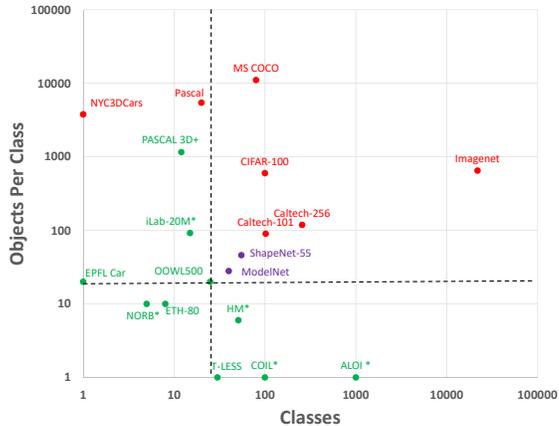}
  \caption{Scatter plot of datasets listed in 
    Table~\ref{tab:dataset}. In the wild datasets are shown in red, in the
    lab in green, and synthetic in purple. An asterisk indicates that
    turntables were used for data collection.}
  \label{fig:scatter}
  \vspace{-10pt}
\end{wrapfigure}

\vspace*{-\vs pt}
\subsubsection{In the lab datasets}
The study of these questions has motivated several ``in the lab'' datasets,
annotated for factors like pose, distance, or lighting. In fact, as shown in
the second third of Table~\ref{tab:dataset}, there is
a long history of such datasets.
However, most of them are below the $60,000$ image threshold, and  too small to train deep CNNs. 
Furthermore, they all have significant
differences compared to OOWL500. First, they exhibit
a professional photography bias similar to that of in the wild datasets. Sometimes the bias is
even stronger, because images are collected in a precisely controlled studio environment; they do not exhibit the camera shake found in OOWL500.
Second, objects are often carefully selected from 
fine-grained classes, such as faces or toys.
Due to this, many of these
datasets have few classes of overlap with in the wild datasets like ImageNet
(only HM comes close to OOWL500 in this regard). This prevents studies of
robustness, such as the ones we pursue in this work, or the use of these
datasets to augment ImageNet. Third, beyond total number of images,
an object dataset should contain a number of classes sufficiently
large to pose a challenging discrimination problem, and a number of object
instances large enough to capture the diversity of appearances of each
class. Even the largest in the lab datasets can be small in this respect.
For example, while iLab-20M contains many images, it only includes $15$ classes
of toy vehicles. This is a relatively small classification challenge.
On the other hand, HM contains $51$ classes, but only a few objects
($6$ on average) per class. This is not enough to capture the appearance
diversity of most object classes. Since dataset size
is combinatorial on these factors, any dataset must achieve a good compromise
between them. Fig.~\ref{fig:scatter} visualizes this trade-off,
by showing the number of classes vs. objects per class of each dataset
in Table~\ref{tab:dataset}. Among in the lab datasets (shown
in green), OOWL500 achieves the best trade-off between the two factors.
Naturally, in the wild datasets tend to be larger in size. However, the scalability of the drone-based data collection process
used by OOWL500, together with the potential for community wide collaborative
collection, could push it to the red area of the plot. It is hard to see
this happening with turntable or dome-based methods.

\vspace*{-\vs pt}
\subsubsection{Synthetic datasets} Recently, there has also been interest in
large datasets of object shape. These are usually composed of synthetic
images, rendered from 3D graphics models. Two popular datasets in this
class, ModelNet~\cite{wu20153d} and ShapeNet~\cite{shapenet2015} are listed in the
bottom third of Table~\ref{tab:dataset}. While object views are comprehensive,  Fig.~\ref{fig:scatter} shows that
they are not much larger than OOWL500 in terms of object classes or
objects per class. Furthermore, these datasets have been primarily used for
operations such as shape classification or retrieval on synthetic
images. It is not clear whether they can be leveraged to improve
real-world object recognition systems. We study this question
in Section~\ref{Guidelines for data collection}.

\vspace*{-\vs pt}
\section{Experiments}\label{experiments}
\vspace*{-\vs pt}
In this section, we report various experiments using OOWL500.
\vspace*{-\vs pt}
\subsection{Experimental setup}\label{subsec:experimental setup}
\vspace*{-\vs pt}
\subsubsection{Datasets} The goal of these experiments were to compare
in the wild,  in the lab, and synthetic datasets, represented
by ImageNet, OOWL500, and ModelNet respectively. Given a 3D shape from ModelNet, 2D images were rendered as in~\cite{su2015multi}. A preliminary step was used to generate three
datasets of comparable classes. Since ImageNet is more fine-grained than
the others, several of its classes were merged in a larger class with the same
number of images. For example,  ``airliner'' and ``warplane'' were merged into an ``airplane'' class. The $1,300$ training and $50$
test images of each original class were randomly sampled ($650$ training, $25$
test images per class) to produce a $1,300$ ($50$) image ``airplane''
training (test) set. Appendix A and B present a list of the merged
ImageNet sub-classes for each object class also found on OOWL500 and ModelNet, respectively.
After this fusion, there was an overlap of $23$ classes between ImageNet and OOWL500,
and $8$ between ImageNet, OOWL500, and ModelNet. To maximize the number of classes per experiment, several new
datasets were created. {\it IN}(23) contains the ImageNet images of the
$23$ classes shared with OOWL500,
 {\it IN}(8) of the $8$ classes
shared by all 3 datasets, and {\it OOWL500}(8) the same for OOWL500.
All experiments were performed with AlexNet~\cite{krizhevsky2009learning},
ResNet~\cite{he2016deep}, and VGG~\cite{simonyan2014very}. Models pre-trained on the full
ImageNet and fine-tuned on each of the IN datasets are denoted
{\it ImageNet classifiers (INC)\/}. For example, {\it INC(23)\/} indicates the
classifiers fine-tuned on IN(23). Models fine-tuned on OOWL500 and ModelNet
are denoted {\it OC\/} and {\it MNC\/} classifiers, respectively.

\vspace*{-\vs pt}
\subsubsection{Frontal Pose} \label{subsec:Frontal Pose}
Comparing recognition accuracy as a function of pose requires a pose
reference, i.e. the $0^o$ view or {\it frontal pose\/}.
This is hard to define for classes like airplanes.
The most natural frontal poses (the plane's nose
or side) are also uncommon on ImageNet, where airplanes are usually seen at $45$
degrees. Because of this, INCs have different ``favorite poses''
for different classes, under the OOWL500 definition of frontal pose.
To avoid bias towards OOWL500, we adopted
a data-driven definition instead. In all experiments, {\it frontal pose\/}  
is the view of largest average recognition
accuracy by the INC classifiers for each class.

\vspace*{-\vs pt}
\subsection{ImageNet bias} \label{subsec:ImageNet bias}

We now report on experiments testing the usefulness
of OOWL500 as a tool to 1) identify biases of ``in the wild''
data collection and 2) help correct them.

\vspace*{-\vs pt}
\subsubsection{Adversarial attacks} \label{subsec:Adversarial attacks}

\begin{figure}[t]\RawFloats

  \begin{minipage}{0.48\linewidth}
    \centering
      \includegraphics[width=\textwidth]{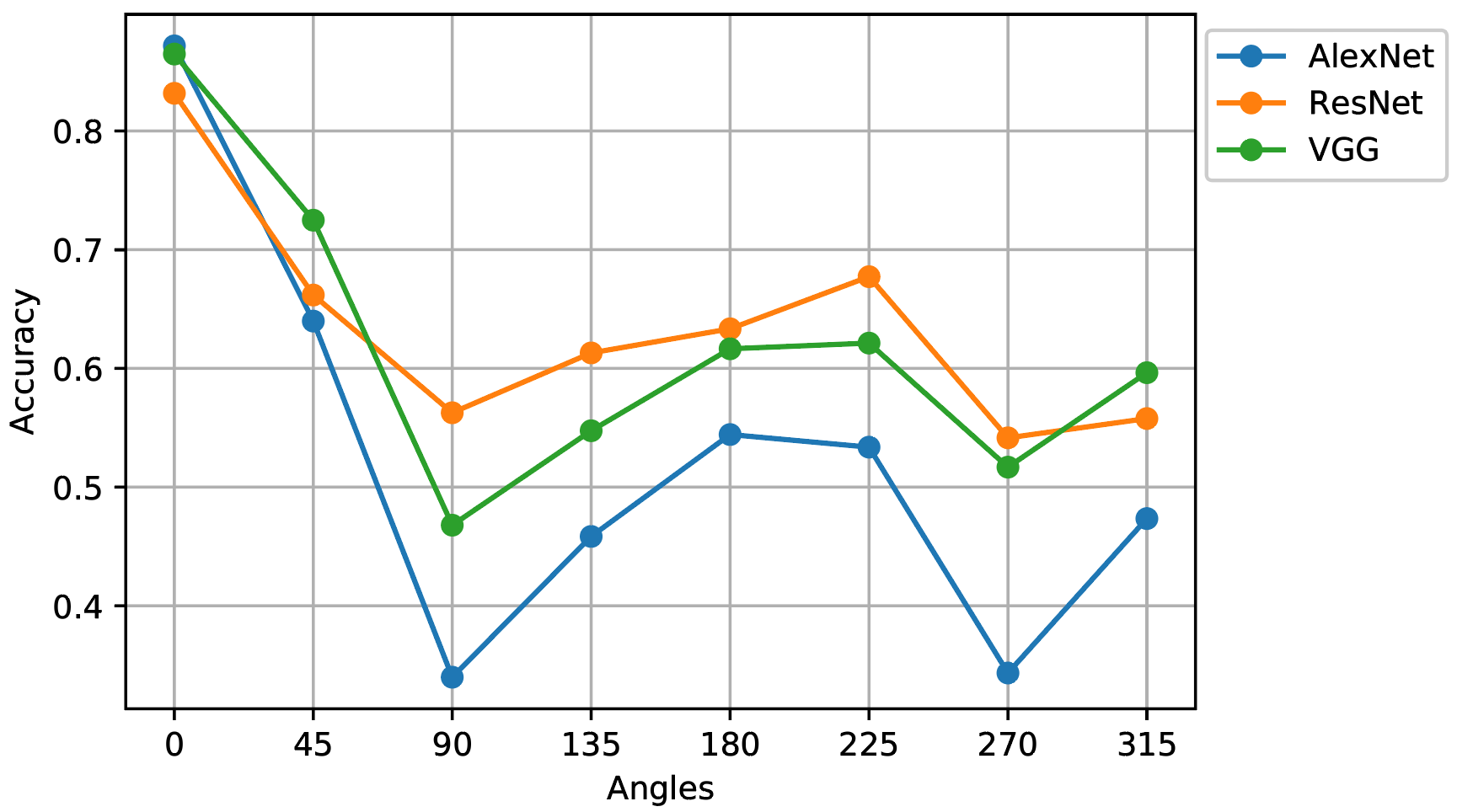}
      \caption{Recognition rate as a function of angle perturbation $\delta \theta$. $\delta \theta=0$ reports to camera shake, $\delta \theta > 0$ to an angle change of $\delta \theta$.}
      \label{fig:attacks}
  \end{minipage}%
  \hspace{8pt}
  \begin{minipage}{.48\linewidth}
    \centering
    \includegraphics[width=\textwidth]{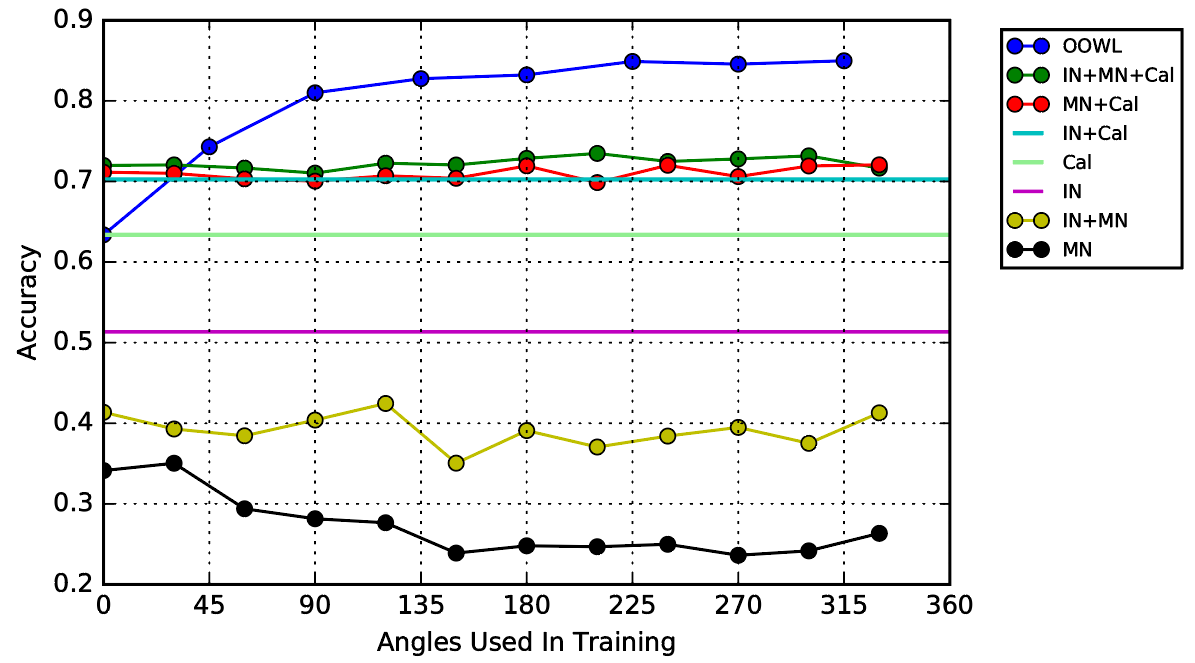}
    \caption{Test set accuracy, on OOWL500, of classifiers trained on 8 training sets, as a function of the pose angles included in the training set. }
    \label{fig:Train_with_synthetic_data}
  \end{minipage}
\end{figure}

There has recently been much interest on adversarial attacks on CNNs~\cite{papernot2017cleverhans,kurakin2016adversarial}. These consist of perturbing an image in a way that
does not affect human recognition and measuring the drop of recognition
accuracy by the CNN. We considered an attack strategy on the INC(23)
classifiers using OOWL500 images. A random frontal pose of OOWL500,
correctly classified by the CNN, was perturbed into another image of
either the same or a different pose. Images of the same pose differ by
camera shake (small perturbation), while those of different poses differ by
view and camera shake (larger). The difference of view angles is
denoted the angle perturbation $\delta \theta$. This was repeated for all
classes of OOWL500. Fig.~\ref{fig:attacks} shows the recognition
rates of the three classifiers as a function of $\delta \theta$. 
The attacks are quite successful, producing a low recognition rate
for most values of $\delta \theta$. They exhibit the  two
ImageNet biases' effects illustrated in Fig.~\ref{fig:egimages}.
The drop to $80-90\%$ recognition rate at $\delta \theta = 0$, 
where all perturbations are due to camera shake, is
explained by the ``professional photography'' bias of ImageNet.
The even lower recognition rates for larger $\delta \theta$
are explained by its ``pose bias''. Note that the latter is
particularly strong, with a performance drop larger than $50$ points
between frontal and least favorite pose!

      


\vspace*{-\vs pt}
\subsubsection{Correcting ImageNet biases} \label{subsec:Correcting ImageNet biases}
We next studied the usefulness of ``in the lab'' datasets such as OOWL500 to
correct ImageNet biases. A first set of experiments investigated whether
OOWL500 is a useful complement to ImageNet. These were inspired
by previous claims of superhuman performance for the classification of
ImageNet object classes. While this has been shown to hold when
ImageNet classifiers are evaluated on the ImageNet test set~\cite{Pneumonia2017paper,HeZR015,AIindex2017report}, it has not
been shown how their generalization to a more adversarial set up compares
to that of humans. To evaluate this, we measured the recognition accuracy of the INC(23)
classifiers on IN(23) and OOWL500, and compared them to those of OOWL
classifiers and human performance. Human performances were evaluated by taking the majority vote of 3 individual Turkers, asked to classify test set images from OOWL500 and
ImageNet. Since all classifiers behaved similarly,
we only report the average across the AlexNet, VGG, and ResNet architectures,
shown in Table~\ref{tab:bias}. The first striking observation is that
the INC(23) classifiers performed quite poorly on OOWL500. While their
performance on the IN(23) test set is only $9$ points weaker than that of
Turk workers, the difference on OOWL500 is $46$ points! 
Note that the differences between OOWL500 and ImageNet pose no difficulties to humans, which have similar performance on both datasets. This demonstrates the vulnerability of ImageNet classifiers to camera shake and pose bias, the importance of datasets (such as OOWL500) that account for these factors, and the fallacy of claims of superhuman performance, even on ImageNet classes.


\setlength\intextsep{10pt}
\begin{wraptable}{r}{4.5cm}
    \caption{Average recognition accuracy of AlexNet, VGG, and ResNet classifiers fine-tunned on IN(23) (denoted INC(23)), OOWL500 (denoted OC), and both (denoted both), on the test sets of IN(23), OOWL500, and ALOI. Experiments on IN(23) and OOWL500 are based on 23 classes, while those on ALOI are based on 13 classes. Human performance
was measured by majority vote among three Turk workers.
}
    \begin{scriptsize}
      \centering
    \begin{tabular}{|c|c|c|c|c|}
      \hline
      & \multicolumn{3}{c|}{Dataset} \\
      \hline
      Classifier& IN(23) & OOWL500 & ALOI \\
     \hline
      INC & 0.79 & 0.33 & 0.55 \\
      OC & 0.44 & 0.77 & 0.71 \\    
      Both & 0.76 & 0.78 & 0.72 \\
      \hline
      Human & 0.88 & 0.87 & - \\
      \hline    
    \end{tabular}
    \end{scriptsize}
    \label{tab:bias}
    \vspace{-5pt}
\end{wraptable}

Note that the classifier
architectures are perfectly capable of handling OOWL500. When
fine-tuned on the latter, the (OC) classifiers achieve good performance
on this dataset. However, there is a significant
drop of accuracy on IN(23). This is a well known problem,
wherein fine-tuning causes the classifier to ``forget'' the
original dataset~\cite{KirkpatrickPRVD16,french1999catastrophic}. These results show that OOWL500 is a good
testing ground for future research in this area. An interesting
additional observation, perhaps the most surprising of this experiment,
is that the OC classifiers, fine-tuned to OOWL500, generalized better
to IN(23) -- $44\%$ accuracy -- than the INC(23), fine-tuned to IN(23),
generalized to OOWL500 -- $33\%$. Finally, the classifiers fine-tuned
on {\it both\/} IN(23) and OOWL500 nearly matched the accuracy of the
classifiers fine-tuned to each dataset. While this was expected, it shows
that {\it complementing\/} ImageNet with OOWL500 has no degradation on the
ImageNet domain, but can substantially increase performance beyond it.

Overall, this experiment supports several conclusions.
First, claims of superhuman performance regarding deep learning classifiers
trained on ImageNet are widely exaggerated. It is clear that the
classifiers do not even generalize for a dataset with {\it the same\/}
object classes, when the images of the latter are not collected on the web.
Second, the two datasets are clearly {\it complimentary.\/} When
classifiers are fine-tuned on both, overall recognition performance is
far superior than when they are fine-tuned on only one. Third, the strongest
transfer from OOWL500 to IN(23) rather than vice versa suggests that there is more
than a simple mismatch of image statistics between the two datasets.
Instead, it suggests the existence of an ImageNet pose bias which hurts
the ability of ImageNet classifiers to generalize to data with the pose
diversity of OOWL500.

The last point is somewhat muddled by the fact that the classifiers
(OC) fine-tuned on OOWL500 were initially pre-trained on ImageNet.
Their better generalization to IN(23) could be due to the pre-training.
To test this, the performances of INC and OC were evaluated on a third dataset ALOI~\cite{geusebroek2005amsterdam} which exhibits significant pose variability and is completely
independent of IN and OOWL500. Since it shares 13 classes with IN and OOWL500, {\it IN}(13) and {\it OOWL500}(13) were created with the procedure of Section ~\ref{subsec:experimental setup}. OC(13) and IN(13) classifiers, fine-tunned on these datasets, were finally tested on ALOI. As shown in Table~\ref{tab:bias},
OC(13) generalized significantly better than INC(13), achieving a gain
of $16$ points and performance nearly identical to that of the classifier
fine-tuned on {\it both\/} IN(13) and OOWL500. This confirms
the need for in the lab datasets that, like OOWL500, enable 
training that specifically corrects ImageNet biases.
\vspace*{-\vs pt}
\subsection{Guidelines for data collection} \label{Guidelines for data collection}

So far, we have seen that OOWL500 can be used to 1) construct adversarial
attacks on and 2) improve generalization performance of ImageNet classifiers.
The final set of experiments exploited OOWL500 to obtain data collection insights.
\vspace*{-\vs pt}
\subsubsection{Alternative training strategies}

Computer generated synthetic datasets provide a simple alternative to ``in the lab'' datasets.
In fact, recent works use synthetic data to improve the generalization of
classifiers~\cite{xiang2015data} or solve problems such as shape retrieval~\cite{su2015multi,bai2016gift}.
To evaluate the potential of synthetic data augmentation,  we fine-tuned 
ImageNet classifiers on the combined training sets of IN(8) and
MN and tested their recognition performance on OOWL500(8).
To characterize sensitivity to pose, various experiments
were performed. Starting from IN(8), all the images with
the same pose of MN were gradually added to the fine-tuning set.
The first experiment included only frontal poses, then $45^o$ views
were added, and so forth. The recognition performance was recorded
after each new pose set was added and is displayed in a cumulative
plot in Fig.~\ref{fig:Train_with_synthetic_data}.
Various transfer modes were compared. Modes IN and MN are baselines using
ImageNet and ModelNet fine-tuning only, respectively. Mode IN+MN used the
combination of the two datasets. To minimize the difficulty of 
transfer due to differences in the overall statistics
of OOWL500, ImageNet, and ModelNet, we also considered a few ``calibrated''
modes where the fine-tuning set included the frontal pose of OOWL500.
Mode Cal used only this set of images, modes IN-Cal and MN-Cal combined the
IN and MN sets with the calibration set respectively, and mode IN-MN-Cal
used all fine-tuning modes. All these transfer modes were compared to
transfer based on fine-tuning on the OOWL500 training set. 

Several conclusions can be made. First, synthetic data does {\it not\/} help
the transfer at all. MN had very poor performance, which in fact
decreased as poses were added. The addition of MN to IN also decreased
the performance of the latter. Second, the biggest gains came from the
calibration modes. The addition of the OOWL500 frontal pose significantly
improved on all the other configurations. While this was somewhat expected,
the gain is quite significant -- at least $20$ points. Again, the addition
of MN poses had no noticeable increase in performance. Third, the
best performance was obtained with OOWL500 alone. While the addition of
IN(8) and MN improved on OOWL500 fine-tuning with the frontal pose only,
there were {\it no gains\/} after that. Adding a {\it single\/} pose
($45^o$) of OOWL500 to the fine-tuning set was enough to match the performance
of adding the entire IN(8) and MN training sets! Adding one further pose
($90^o$) raised the recognition rate to a level close to the overall best.
In summary, these experiments show that 1) synthetic augmentation {\it is
not the solution\/} to the problem of ImageNet biases and 2) even
a small number of in the lab poses guarantee major gains over the
ImageNet performance.

\begin{figure}[t]\RawFloats
  \centering
    \begin{tabular}{cc}
    \includegraphics[width=.45\textwidth]{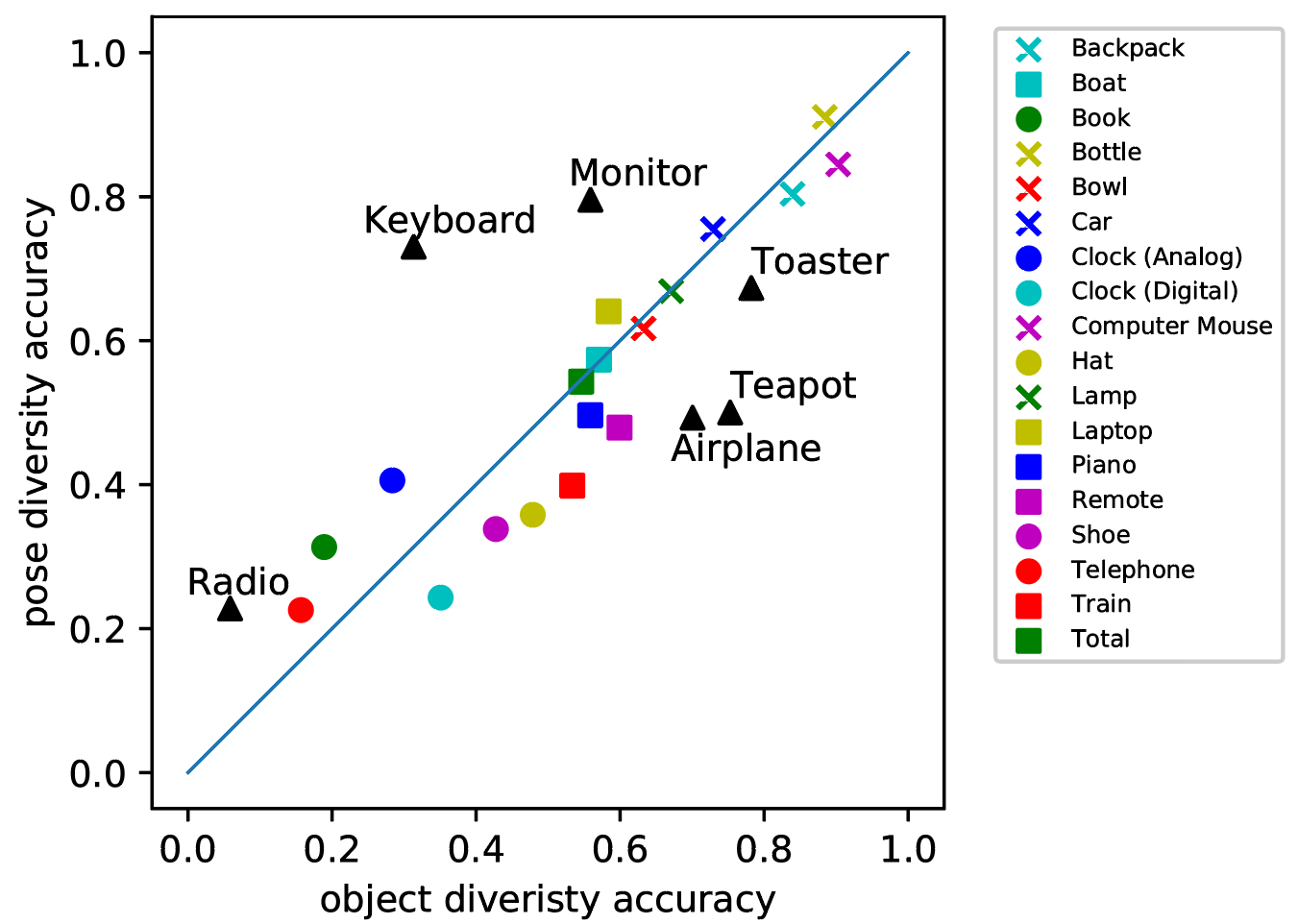} &
    \includegraphics[width=.45\textwidth]{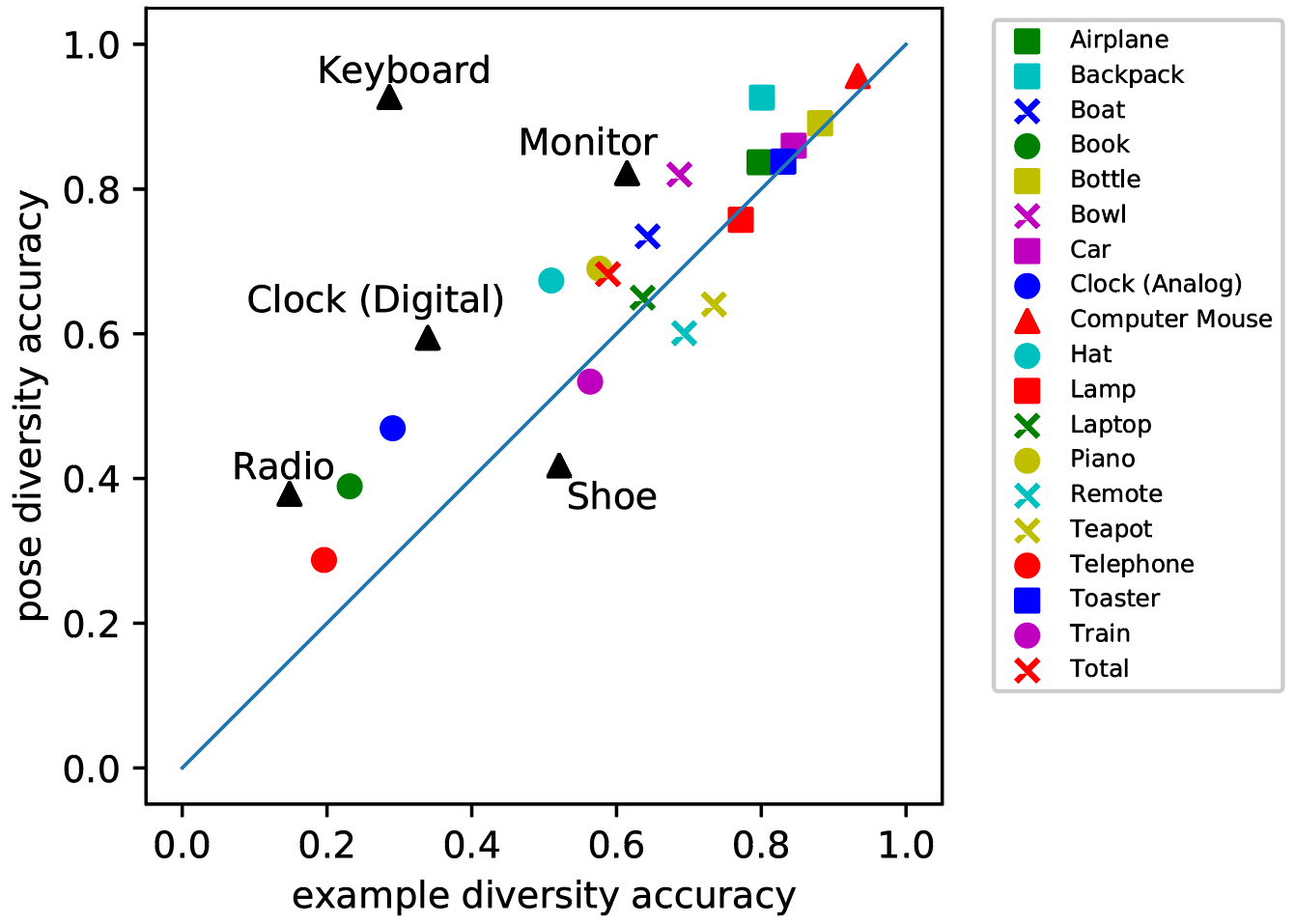} \\
    \scriptsize{(a)} & \scriptsize{(b)}
  \end{tabular}
  \caption{Recognition accuracy vs. dataset parameters $O/C,P/O,E/P$ for a fixed total training set size ($D$). (a) x axis shows accuracy by training with $(O/C,P/O) = (16,1)$, which favors object diversity; y axis by training with $(2,8)$, favoring pose diversity. (b) x axis shows accuracy by training with $(E/P,P/O) = (30,1)$, favoring camera shake; y axis by training with $(3,8)$, favoring pose diversity.}
  \label{fig:fxd}
\end{figure}

\vspace*{-9 pt}
\subsubsection{Pose diversity vs object diversity}

Object recognition has many dimensions of image variability, such as the
diversity of pose or appearance of the objects in each class. For a fixed
dataset size, it is always necessary to establish trade-offs for
the coverage of all these dimensions. In the lab datasets
such as OOWL500 enable the study of how this coverage impacts the quality of
the resulting classifiers. OOWL500 contains four such dimensions:
the number of classes ($C$), the number of object instances per
class ($O/C$), the number of poses per object ($P/O$), and the number of
examples per pose of each object ($E/P$). For a fixed dataset size
$D = C \times O/C \times P/O \times E/P$, many configurations of
these factors are possible. For example, ImageNet uses $E/P = P/O = 1$ and
favors {\it object diversity,\/} i.e. large $O/C$. OOWL500 enables other
strategies, such as {\it pose diversity\/} (larger $P/O$) or
{\it camera shake\/} (larger $E/P$). Two experiments were performed to
compare the relative performance of such strategies. In both cases,
we fixed the number of classes and training set size ($C=23$ and $D=11040$
respectively). The first experiment compared the importance of object vs. pose
diversity. For this, we fixed  $E/P = 30$ and considered two extreme
configurations. The first, denoted {\it max object diversity\/}, used a
fine-tuning set with the largest possible $O/C = 16$ and smallest $P/O=1$.
The second, denoted {\it max pose diversity\/}, used the smallest $O/C = 2$
and the largest $P/O = 8$. The recognition accuracy of classifiers
trained with these fine-tuning sets was then compared.

The scatter plot of Fig.~\ref{fig:fxd}(a) illustrates the average recognition
accuracies  of the two strategies for all object classes
in OOWL500. Points on the blue line signal classes with
equal recognition rate under the two strategies. Points above (below)
signal classes that are favored by max pose (object) diversity. 
Interestingly, more than half the classes have no preference or {\it prefer pose diversity.\/} While this was expected for some classes such as ``Monitor'' or ``Keyboard,'' which have relatively low object diversity but whose appearance can vary significantly with pose, it was surprising that it holds for so many. Note that most in the wild datasets only exhibit object diversity, because
pose diversity is almost impossible to control in this setting. 
The fact that the preference for object or pose diversity varies with the class also suggests that further research is needed on intelligent data collection strategies that can take this into account. In the lab datasets, like OOWL500, enable such research. 

\vspace*{-\vs pt}
\subsubsection{Pose diversity vs camera shake}
A second experiment compared the importance of pose diversity vs.
camera shake. For this, we fixed  $O/C = 16$ and considered two
strategies to assemble the fine-tuning set. {\it Max pose diversity\/}
used $P/O=8$ and $E/P=3$; {\it max camera shake\/} used $P/O = 1$
and $E/P=30$. The scatter plot in Fig.~\ref{fig:fxd}(b) shows the average
recognition accuracies results. Points above (below)
the blue line signal classes that are favored by max pose diversity (camera
shake). Most classes have a clear {\it preference to pose diversity,\/} an effect even stronger than in the previous experiment. Again, while camera shake is usually simulated for (by shifting, cropping, etc.) in classifier design, pose diversity is much harder to address with in the wild datasets. 


%

\vspace*{-\vs pt}
\section{Conclusion}
\vspace*{-\vs pt}
In this work, we have investigated the hypothesis that dataset collection
``in the wild'' is not sufficient to guarantee unbiased object recognizers.
 Specifically, we have argued that ``professional photography''
and ``pose'' biases are difficult to eliminate in web-based data collection
and proposed a new ``in the lab'' data collection setup that mitigates
these factors. This consists of using a drone to take 
images as it hovers in various spots around an
object. We have shown that this setup has multiple interesting
properties. First, by utilizing drone flight, 
it is possible to collect images of objects at controlled camera angles. This
can be used to compensate pose bias. Second,
the procedure is inexpensive and easily replicable. Hence, it could
lead to a scalable data collection effort by the vision community.
Third, drone captured images tend to have a certain amount of camera
shake that mitigates professional photography bias.

The usefulness of the procedure was demonstrated through the  collection
of the OOWL500 dataset -- the largest in the lab dataset recognition
dataset in the literature when both number of classes and objects per
class are considered. OOWL500 was then shown to enable a number of new
insights on object recognition. 
First, a novel adversarial attack strategy was developed,
where image perturbations are defined in terms of semantic properties
such as camera shake and pose. Experiments with this strategy have shown
that ImageNet indeed has considerable amounts of pose and professional
photography bias. Second, it was used to show that the augmentation of
in the wild datasets, such as ImageNet, with in the lab data, such as
OOWL500, can significantly decrease these biases, leading to
object recognizers of improved generalization. Third, it was used
to study various questions on ``best procedures'' for dataset collection.
This has shown that data augmentation with synthetic images does not suffice
to eliminate in the wild datasets biases, and that camera shake
and pose diversity play a more important role in object
recognizers' robustness than previously thought.

\bibliographystyle{splncs}
\bibliography{posebib}

\end{document}